# Shallow Network Based on Depthwise Over-Parameterized Convolution for Hyperspectral Image Classification

Hongmin Gao, *Member, IEEE,* Zhonghao Chen, *Student Member, IEEE,* and Chenming Li

*Abstract*—**Recently, convolutional neural network (CNN) techniques have gained popularity as a tool for hyperspectral image classification (HSIC). To improve the feature extraction efficiency of HSIC under the condition of limited samples, the current methods generally use deep models with plenty of layers. However, deep network models are prone to overfitting and gradient vanishing problems when samples are limited. In addition, the spatial resolution decreases severely with deeper depth, which is very detrimental to spatial edge feature extraction. Therefore, this letter proposes a shallow model for HSIC, which is called depthwise over-parameterized convolutional neural network (DOCNN). To ensure the effective extraction of the shallow model, the depthwise over-parameterized convolution (DO-Conv) kernel is introduced to extract the discriminative features. The depthwise over-parameterized Convolution kernel is composed of a standard convolution kernel and a depthwise convolution kernel, which can extract the spatial feature of the different channels individually and fuse the spatial features of the whole channels simultaneously. Moreover, to further reduce the loss of spatial edge features due to the convolution operation, a dense residual connection (DRC) structure is proposed to apply to the feature extraction part of the whole network. Experimental results obtained from three benchmark data sets show that the proposed method outperforms other state-of-the-art methods in terms of classification accuracy and computational efficiency.**

## I. INTRODUCTION

Hyperspectral images (HSIs) can reflect hundreds of bands of imaging information of an object in the visible to near-infrared band, containing not only the object's geometric structure information in the spatial dimension, but also the object's rich continuous spectral information in the spectral dimension [1]. Based on this fact, it provides important data support to predict the category of each pixel of the imaging object. In recent decades, hyperspectral image classification (HSIC) technology has been extensively researched and implemented in a variety of fields [2]. Meanwhile, plenty of classification algorithms have been developed.

In the early research of HSIC, a series of traditional machine learning-based classification methods (e.g., support vector machine (SVM) [3], maximum likelihood [4]) have been proposed and demonstrated good classification results. However, the limited labeled samples and the Hughes phenomenon [5] caused by the high-dimensional characteristics of HSIs will seriously limit the classification accuracy of these methods. In general, the manual features derived by early traditional classification methods have weak representation capacity, which cannot obtain satisfactory classification performance, especially poor generalization.

Recently, the excellent feature extraction performance of convolutional neural network (CNN) has attracted much attention in a variety of fields. Certainly, in the field of HSIC, a great number of CNN-based models have been developed to improve HSIC performance [6] - [12]. For example, Chen et al. [7] designed a deep network based on 3D-CNN to realize the end-to-end classification of hyperspectral images. However, as it is a simple concatenation of a few conventional convolutional layers, it does not extract enough features for classification. Therefore, to extract features with more abstract representative ability, the deeper networks [9] - [10] have been investigated widely. However, the deep network structure will lead to the gradient vanishing and poor model fitting. In this regard, residual network (ResNet) [14] and dense network strategy (DenseNet) [15] are applied to many deep models [11] - [13], to alleviate the problem of gradient vanishing problem. HSIs, on the other hand, have more channels than RGB images, resulting in a high number of training parameters when employing a conventional 2D convolution kernel [16]. Therefore, many methods [16] - [17] apply depthwise convolution [18] to HSIC to make the model lighter. However, another problem brought by the deep convolution neural network model is that the resolution of the input sample image will decrease with the increase of convolution times, which is very unfavorable to the extraction of spatial edge features [19], especially in the case of low spatial resolution of hyperspectral images. Although dilated convolution [20] does not diminish spatial resolution, the gridding problem that results lose spatial feature continuity. Therefore, it is meaningful to study the efficient classification of hyperspectral images under the condition of using as few convolution layers as possible to reduce the model depth. Furthermore, the shallow structure of

This work was supported in part by the National Natural Science Foundation of China under Grant 62071168. in part by the National Key Research and Development Program of China under Grant 2018YFC1508106, and in part by the Fundamental Research Funds for the Central Universities of China under Grant B200202183. *(Corresponding author: Chenming Li.)*

Hongmin Gao, Zhonghao Chen, and Chenming Li are with the College of Computer and Information, Hohai University, Nanjing 211100, China (e-mail: gaohongmin@hhu.edu.cn; chenzhonghao@hhu.edu.cn; lcm@hhu.edu.cn).

the network will significantly reduce the model parameters and thus make the network more lightweight.

In this letter, a novel shallow model for HSIC based on depthwise over-parameterized convolution (DO-Conv) [21] is proposed. Different from the aforementioned convolutional layers used in methods, DO-Conv combines standard convolution kernel with depthwise convolution kernel to form a new convolution kernel structure, which can enhance the ability of feature extraction of one layer. In addition, in order to improve the utilization efficiency of features, inspired by [13] - [14], the densely residual connection (DRC) structure is implemented to the model to improve the classification performance of the shallow network model by reducing feature loss.

The main contributions of this letter are summarized as follows:

1） A lightweight shallow network structure with only two feature extraction layers is proposed for HSIC.

2） In order to improve the capability of feature extraction, DO-Conv is investigated for HSIC for the first time.

3） In addition, the proposed DRC structure reduces the loss of low-level edge information at the output by fusing the high-resolution spatial edge features before the convolution operation with the semantic features extracted by each subsequent convolution.

## II. METHODOLOGY

DO-Conv and DRC structure is the core of the DOCNN-DRC. In this section, we first give a detailed introduction of the architecture of DO-Conv, then discuss the advantage of DRC structure in the task of HSIC, and finally, we describe the overall framework of DOCNN-DRC.

### A. Structure of DO-Conv

*1) Standard Convolution and Depthwise Convolution:* Convolution operation, as we all know, involves sliding the convolution kernel on the input image and calculating the grey value of the associated image pixel in the convolution kernel center to complete the entire image feature extraction. For the convenience of understanding, we set the input image as a 2D tensor $P \in \mathbb{R}^{(W \times H) \times C_{in}}$, Where $W$ and $H$ are the spatial dimensions of the input image, $C_{in}$ indicates the number of channels of the input image. Specifically, the convolution operation of standard convolution is performed simultaneously on all channels of the image, with the weight being shared. As depicted in Fig. 1 (a), the convolution kernel of standard convolution can be set as $W \in \mathbb{R}^{C_{out} \times (W \times H) \times C_{in}}$, where $C_{out}$ indicates the number of channels output. Therefore, the standard convolution operation $\star$ can be formulated as $O = W \star P$:

$$O_{c_{out}} = \sum_{i}^{(W \times H) \times C_{in}} W_{c_{out} i} P_i \qquad (1)$$

where $O \in R^{C_{out}}$ denotes the output result (for the convenience of display, the padding operation is not applied to all convolution operations in this letter, and use bias are not considered). However, the parameter sharing of different channels will make some channel sensitive feature extraction inadequate, especially when we want to extract various features of different channels. This is notably essential in the feature extraction of HSIs with a lot of channels. The depthwise convolution, on the other hand, can execute individual convolution calculations for each channel of the input image, taking into consideration each channel's unique spatial properties. As shown in Fig. 1 (b), the convolution kernel of depthwise convolution can be set as $W \in \mathbb{R}^{(W \times H) \times D_{mul} \times C_{in}}$, where $D_{mul}$ is the depth multiplier. Therefore, each $W \times H$ -dimensional feature can be transformed to a $D_{mul}$-dimensional feature. Accordingly, the depthwise convolution operation $\circ$ can be expressed as $O = W \circ P$:

$$O_{d_{mul} c_{in}} = \sum_{i}^{W \times H} W_{i d_{mul} c_{in}} P_{i c_{in}} \qquad (2)$$

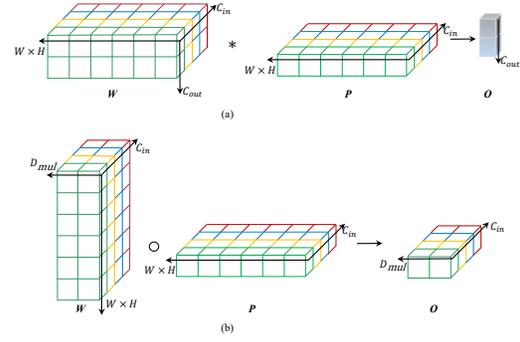

Fig. 1. (a) Architecture of standard convolution. (b) Architecture of depthwise convolution.

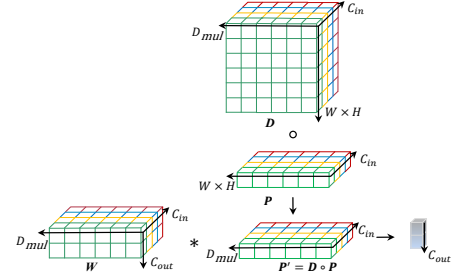

Fig. 2. Architecture of DO-Conv

It is easy to observe that each element in $O \in R^{D_{mul} \times C_{in}}$ is obtained by multiplying the vertical column of W by each channel point of $P$. However, it is obvious that the independent depthwise convolution ignores the correlation of channel features.

*2) DO-Conv:* Compared with the image in the computer vision task, HSIC task has smaller sample patch size and data amount. Therefore, for limited training samples, it is essential to obtain more adequate feature extraction. As mentioned in the previous section, standard convolution and depthwise convolution are complementary in feature extraction. As a result, combining the properties of the two convolution kernels to create a new convolution kernel is a plausible option. DO-Conv is a novel convolution kernel structure that combines the two convolutions mentioned previously. When compared to single depthwise convolution, it is clear that DO-Conv will raise a certain amount of training parameters. As a result, this can be thought of as an instance of depthwise convolution over parameterization. The transformation process of the feature map in DO-Conv is shown in Fig. 2. It can be seen that the DO-Conv operation is composed of a depthwise convolution and a standard convolution. The convolution kernel of depthwise convolution and standard convolution can be set as $\boldsymbol{D} \in \mathbb{R}^{(W \times H) \times D_{mul} \times C_{in}}$, and $\boldsymbol{W} \in \mathbb{R}^{C_{out} \times D_{mul} \times C_{in}}$. Therefore, for a given input image $\boldsymbol{P} \in \mathbb{R}^{(W \times H) \times C_{in}}$, after the DO-Conv operation, we get the feature output $\boldsymbol{O} = (\boldsymbol{D}, \boldsymbol{W}) \odot \boldsymbol{P}$, $\boldsymbol{O} \in \mathbb{R}^{C_{out}}$. More specifically, the DO-Conv operation's feature map transformation process can be represented as the corresponding formula:

$$\boldsymbol{O} = (\boldsymbol{D}, \boldsymbol{W}) \odot \boldsymbol{P}$$
$$= \boldsymbol{W} * (\boldsymbol{D} \circ \boldsymbol{P}). \qquad (3)$$

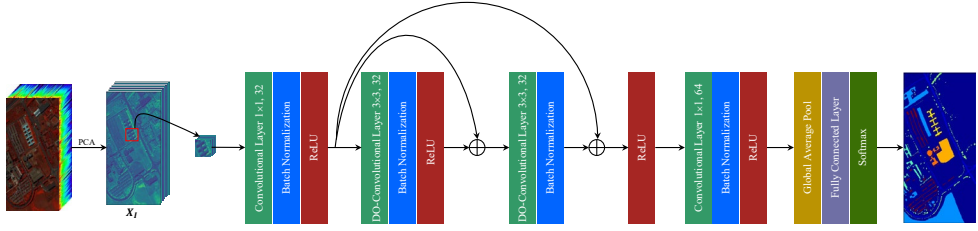

Fig. 3. Schematic of the proposed DOCNN-DRC. ⊕ denotes feature addition.

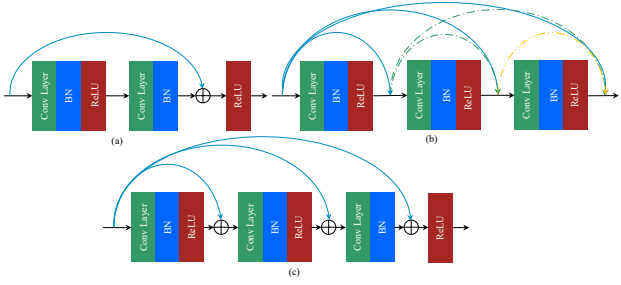

Fig. 4. (a) Architecture of ResNet. (b) Architecture of DenseNet. (c) Architecture of Dense Residual Connection.

Combining formula (3) and Fig. 2, it can be understood that the depthwise convolution first extracts the features of the input image channel by channel and obtains the transformed feature map as $\boldsymbol{P}' = \boldsymbol{D} \circ \boldsymbol{P}$, $\boldsymbol{P}' \in \mathbb{R}^{C_{in} \times D_{mul}}$. Then, the spatial features of $\boldsymbol{P}'$ in the entire channel are extracted using standard convolution, and the output of the feature map is $\boldsymbol{O} = \boldsymbol{W} * \boldsymbol{P}'$, which realizes the fusion of multi-channel features of HSI. In the network structure, the depthwise convolution kernel and the standard convolution kernel are actually integrated into a DO-Conv convolution kernel to complete the convolution calculation process. According to formula (3), the derivation process of the DO-Conv operator $\boldsymbol{Q}$ is as follows:

$$\boldsymbol{O} = (\boldsymbol{D}^T \circ \boldsymbol{W}) * \boldsymbol{P}$$
$$= \boldsymbol{Q} * \boldsymbol{P} \qquad (4)$$

where the $\boldsymbol{D}^T \in \mathbb{R}^{D_{mul} \times (W \times H) \times C_{in}}$ is the transpose of $\boldsymbol{D}$.

### B. Structure of Dense Residual Connection

In the CNN-based HSIC task, gradient vanishing caused by deep network structure is a common problem. It means that when the network parameters will stop updating before the neural network converges. Furthermore, while extracting spatial features, the over-deep network structure would reduce spatial resolution and cause some edge feature loss, which is undesirable for feature extraction of HSIs with low spatial resolution. Literatures have proved that residual network can alleviate the problem of gradient vanishing through a cross-layer direct connection. However, as illustrated in Figure 4 (a), the residual structure is frequently applied locally in the network, which cannot compensate for the loss of spatial information induced by network deepening. Furthermore, as shown in Fig. 4 (b), to further improve feature reuse between layers, DenseNet connects the channel dimensions of feature maps of all layers to their subsequent layers. Although this connection method can ensure that each layer of the deep network contains the shallowest high-resolution features, it will increase the memory consumption and bring redundancy of features. Therefore, inspired by ResNet and DenseNet, a simple method for feature fusion is proposed, which can alleviate the gradient vanishing and avoid the loss of feature. As depicted in Fig. 4 (c), the proposed DRC is presented for feature learning, which reduces the loss of shallow layers' edge features by adding the output of the first layer to the input of all the subsequent layer. In addition, DRC uses less memory overhead than dense connections.

### C. Schematic of the proposed Method

The proposed schematic of HSIC is presented in Fig. 3. First, principal component analysis (PCA) is performed on the original HSIs to generate the reduced dimension data $\boldsymbol{X_I}$. Then, 3-D image cubes are fed into a standard $1 \times 1$ convolutional layer that has 32 filters, which is used to adjust the number of the feature map's channel to facilitate the subsequent feature fusion. Next, a simple feature extraction module combined of two DO-Conv layers and DRC structure is utilized to hierarchically excavate the discriminate features. After that, another standard $1 \times 1$ convolutional layer is employed to boost the feature representation and modify the channels of the feature map from 32 to 64, which also makes features more abundant. Then, a global average pooling layer and a fully connected layer are applied to transform the feature map into the feature vector. Finally, the Softmax function is adopted to obtain the final categorization result.

## III. EXPERIMENTS AND DISCUSSION

### A. Data sets and Experimental Setup

Three famous HSI data sets are used to demonstrate the reliability of the proposed method, which are University of Pavia (PU), Salinas (SA), and Indian Pines (IP). To better test the performance of the model, the number of training samples divided from the dataset is limited. Concretely, for the PU and SA datasets, only 1% of the labeled samples are used to train the model. For the IP dataset, we use 10% of the annotated samples as the training set since there are some categories with a very small number of samples.

TABLE I
CLASSIFICATION ACCURACIES OF DIFFERENT METHODS FOR THE PU DATA SET 1% TRAINING SAMPLES, THE SA DATA SET 1% TRAINING SAMPLES, AND IP DATA SET 10% TRAINING SAMPLES

|  | Class | SVM | 3D-CNN | DCNN | DFFN | SSRN | HybridSN | OURS |
|---|---|---|---|---|---|---|---|---|
| Pavia University | Asphalt | 91.41±1.10 | **99.97±0.03** | 84.99±0.17 | 96.29±2.74 | 99.16±0.14 | 78.66±2.34 | 99.41±0.08 |
|  | Meadows | 97.04±0.65 | 98.23±1.21 | 98.41±0.55 | 99.81±0.07 | 99.86±0.05 | 95.16±1.62 | **99.92±0.04** |
|  | Gravel | 64.82±5.17 | 96.25±0.79 | 96.05±0.53 | 95.39±0.29 | 73.24±1.27 | **97.93±0.86** | 88.74±0.59 |
|  | Trees | 85.53±4.35 | 96.73±0.49 | **98.32±0.23** | 94.61±0.69 | 97.92±0.76 | 93.31±2.57 | 97.46±0.76 |
|  | Painted-M-S | 99.70±0.32 | 99.62±0.27 | **100±0.00** | 99.25±0.09 | 99.55±0.16 | 99.54±0.07 | 99.77±0.27 |
|  | Bare Soil | 69.77±3.50 | 95.36±0.38 | 88.16±5.80 | **100±0.00** | 99.80±0.11 | 99.85±0.19 | **100±0.00** |
|  | Bitumen | 69.02±12.02 | 95.96±0.86 | 88.65±1.39 | **100±0.00** | 96.36±0.69 | 96.00±2.23 | 98.79±0.21 |
|  | Self-B-B | 83.62±3.04 | 66.11±1.22 | 87.62±4.39 | 95.20±0.79 | 95.21±2.31 | **100±0.00** | 96.49±0.76 |
|  | Shadows | 99.57±0.08 | 97.54±0.37 | 93.48±0.89 | **99.36±0.05** | 97.86±1.29 | 84.72±1.47 | 96.79±0.64 |
|  | OA (%) | 88.67±0.34 | 95.14±0.39 | 94.06±0.52 | 97.93±0.36 | 97.73±0.73 | 96.33±0.50 | **98.57±0.23** |
|  | kappa × 100 | 84.75±0.50 | 93.58±0.61 | 92.14±0.89 | 97.25±0.24 | 96.99±1.17 | 95.11±0.67 | **98.10±0.11** |
| Salinas Valley | OA (%) | 87.64±0.29 | 92.46±0.33 | 96.31±0.27 | 97.12±0.29 | 97.64±0.23 | 98.28±0.12 | **98.74±0.13** |
|  | kappa × 100 | 86.19±0.32 | 91.57±0.18 | 96.19±0.31 | 96.79±0.25 | 97.66±0.25 | 98.08±0.13 | **98.6±0.11** |
| Indian Pines | OA (%) | 69.86±2.17 | 92.67±0.36 | 94.76±0.37 | 95.89±0.19 | 98.59±0.38 | 97.99±0.19 | **99.03±0.18** |
|  | kappa × 100 | 64.83±1.94 | 91.63±0.53 | 94.59±0.62 | 97.31±0.27 | 98.39±0.44 | 97.71±0.22 | **98.9±0.15** |

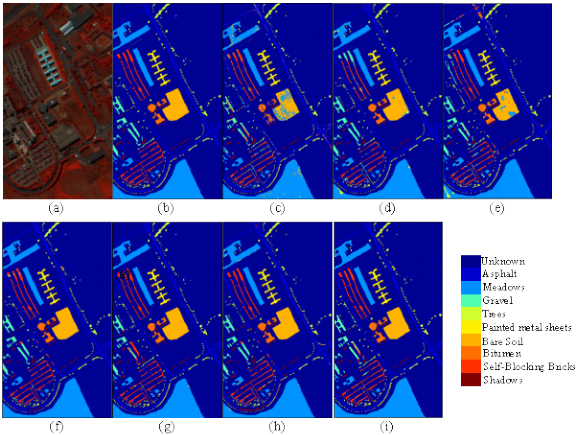

Fig. 5. Classification maps for PU. (a) False-color image. (b) Ground truth (c) - (i) Predicted classification maps for SVM, 3D-CNN, DCNN, DFFN, SSRN, HybirdSN, and DOCNN-DRC.

To evaluate the classification effectiveness of the proposed model, we compared the model with five state-of-the-art algorithms SVM [3], 3D-CNN [5], DCNN [10], DFFN [11], SSRN [12], and HybridSN [22]. For the input of the DOCNN-DRC, the spatial size is set to 9 × 9 with 15 spectral principal components are retained by PCA. The learning rate is configured to 0.01 and the batch size is 64 for all datasets, and the number of training epochs is 120, 120, and 150 for the three datasets, respectively. Furthermore, to guarantee fairness, all parameters of the other comparison methods keep the same settings as those in the corresponding literatures. Moreover, the overall accuracy (OA) and kappa coefficient are utilized as the standard evaluation indicators of classification performance. Especially, all the experimental results are taken as the average of ten runs.

*B. Classification Performance Analysis*

Table I gives the classification accuracy for all methods on the PU, SA, and IP datasets. As reported, the classification performance of the proposed method outperforms the other methods, demonstrating the effectiveness of the proposed shallow structure in extracting discriminative features. Among them, the classification accuracy of traditional machine learning methods is much worse than other deep learning methods. In addition, it is worth noting that SVM shows poor generalization on different data sets. Although 3D-CNN also has a shallow model structure, it does not perform as well with limited samples. However, as an early CNN-based research result, it is a guide to the application of CNN on hyperspectral image classification. Particularly, HybirdSN introduced 2D convolution on top of 3D-CNN, which was efficient in improving the classification accuracy, but the improvement of the classification performance was limited by using only ordinary convolution in shallow models. Compared with the first two shallow models, DCNN, DFFN, and SSRN are proposed as deep models to extract more representative deep abstract features. Accordingly, in order to reduce the gradient vanishing problem caused by model deepening, various residual structures are used. However, although the local residual structure can alleviate the gradient vanishing to a certain extent, the features extracted from the end of the deep model lack high-resolution texture features, which limits the performance of these models.

TABLE II
THE NUMBER OF TRAINABLE PARAMETERS, TRAINING AND TESTING TIMES (IN SECONDS) OF DIFFERENT DEEP MODELS FOR THE PU DATA SET

| Methods | Parameters | Train time | Test time | OA |
|---|---|---|---|---|
| 3D-CNN | 9,867k | 3167.31 | 17.64 | 95.14% |
| DCNN | 545k | 122.99 | 6.83 | 94.06% |
| DFFN | 370k | 2514.63 | 12.41 | 97.93% |
| SSRN | 120k | 111.25 | 12.33 | 97.73% |
| HybridSN | 5,122k | 121.9 | 2.97 | 96.33% |
| OURS | 27k | 5.89 | 1.76 | 98.57% |

More intuitively, the classification map of each approach on the PU data set is depicted in Fig. 5. It can be observed that the classification map of the proposed method is of the highest quality and is most similar to the ground truth.

TABLE III
Effect of DO-Conv Layer and DRC on OA and Kappa on SA and IP Data Sets

|  |  | SCNN | SDCNN | DOCNN | DOCNN-DRC |
|---|---|---|---|---|---|
| SA | OA | 97.80% | 97.89% | 98.41% | 98.74% |
|  | Kappa | 97.19% | 97.36% | 98.04% | 98.62% |
| IP | OA | 96.59% | 97.29% | 98.54% | 99.03% |
|  | Kappa | 96.31% | 97.16% | 98.31% | 98.90% |

To further explicate the computational and storage efficiency of the model, the number of trainable parameters and model execution elapsed time (training and testing) for the PU data set are presented in Table II. Since the model used uses a shallow model design strategy, it has a minimal number of training parameters compared to other methods, which also makes it more efficient to execute. Meanwhile, the proposed model achieves optimal classification performance. It can be stated that the proposed model is competitive in terms of accuracy but also in terms of computing and storage cost when compared to other methods.

### C. Ablation Study

To verify that the DO-Conv and DRC are productive in boosting the classification performance of HSIs, ablation experiments are applied to two data sets. The experiments are carried out on four models: 1). The DRC structure in the original model is removed, and the DO-Conv is replaced by the standard convolution kernel, so the model is called SCNN; 2). The DRC structure in the original model is removed, and the DO-Conv is replaced by the standard depthwise convolution kernel, so the model is called SDCNN; 3). The DRC structure in the original model is removed, and the model is called DOCNN; 4). The model structure proposed in this paper. The corresponding classification results are reported in Table III. Correspondingly, the following conclusions can be summarized: 1). The experimental results of the first three models demonstrate that DO-Conv has superior feature extraction capabilities compared to standard convolution and depthwise convolution, which allows the use of DO-Conv to build shallower models to avoid the problems associated with deeper models and achieve better classification accuracy. 2). The experimental results of the latter two models indicate that the DRC structure can effectively fuse the shallowest features with the subsequent layer features hierarchically to enhance the feature representation ability.

### IV. Conclusion

This letter presented a shallow CNN-based model for HSIC, which is composed of the depthwise over-parameterized convolution kernel and dense residual connection. DO-Conv can extract features more comprehensively for category identification, which can also short the depth of the model structure in HSIC. Furthermore, the DRC is utilized to improve the comprehensiveness of end-of-network features. Extensive experimental results pronounce that our proposed method could achieve promising performance.

### References

[1] L. R. Gao, D. F. Hong, J. Yao et al., "Spectral Superresolution of Multispectral Imagery With Joint Sparse and Low-Rank Learning," IEEE Trans. Geosci. Remote Sens., vol. 59, no. 3, pp. 2269-2280, Mar, 2021.

[2] B. Rasti et al., "Feature Extraction for Hyperspectral Imagery: The Evolution from Shallow to Deep (Overview and Toolbox)," in IEEE Geosci. Remote Sens. Magaz., vol. 8, no. 4, pp. 60-88, Dec. 2020.

[3] F. Melgani and L. Bruzzone, "Classification of hyperspectral remote sensing images with support vector machines," IEEE Trans. Geosci. Remote Sens., vol. 42, no. 8, pp. 1778–1790, Aug. 2004.

[4] J Peng, L. Li, and Y. Y. Tang, "Maximum likelihood estimation-based joint sparse representation for the classification of hyperspectral remote sensing images," IEEE Trans. Neural Netw. Learn. Syst., vol. 30, no. 6, pp. 1790–1802, Jun. 2019.

[5] G. Hughes, "On the mean accuracy of statistical pattern recognizers," IEEE Trans. Inf. Theory, vol. IT-14, no. 1, pp. 55–63, Jan. 1968.

[6] D. Hong, L. Gao, J. Yao, B. Zhang, A. Plaza, and J. Chanussot, ''Graph convolutional networks for hyperspectral image classification,'' IEEE Trans. Geosci. Remote Sens., early access, Aug. 18, 2020.

[7] Y. Chen, H. Jiang, C. Li, X. Jia, and P. Ghamisi, "Deep feature extraction and classification of hyperspectral images based on convolutional neural networks," IEEE Trans. Geosci. Remote Sens., vol. 54, no. 10, pp. 6232–6251, Oct. 2016.

[8] D. F. Hong, L. R. Gao, N. Yokoya et al., "More Diverse Means Better: Multimodal Deep Learning Meets Remote-Sensing Imagery Classification," IEEE Trans. Geosci. Remote Sens., vol. 59, no. 5, pp. 4340-4354, May, 2021.

[9] M. Liang, L. Jiao, S. Yang, F. Liu, B. Hou and H. Chen, "Deep Multiscale Spectral-Spatial Feature Fusion for Hyperspectral Images Classification," in IEEE J. Sel. Topics Appl. Earth Observ. Remote Sens., vol. 11, no. 8, pp. 2911-2924, Aug. 2018.

[10] H. Lee and H. Kwon, "Going deeper with contextual CNN for hyperspectral image classification," IEEE Trans. Image Process., vol. 26, no. 10, pp. 4843–4855, Oct. 2017.

[11] W. Song, S. Li, L. Fang and T. Lu, "Hyperspectral Image Classification with Deep Feature Fusion Network," in IEEE Trans. Geosci. Remote Sens., vol. 56, no. 6, pp. 3173-3184, June 2018.

[12] Z. Zhong, J. Li, Z. Luo and M. Chapman, "Spectral–Spatial Residual Network for Hyperspectral Image Classification: A 3-D Deep Learning Framework," in IEEE Trans. Geosci. Remote Sens., vol. 56, no. 2, pp. 847-858, Feb. 2018.

[13] C. Y. Lin, Z Tao, A. S. Xu et al., "Sequential Dual Attention Network for Rain Streak Removal in a Single Image," IEEE Trans. Image Process., vol. 29, pp. 9250-9265, 2020.

[14] K. He, X. Zhang, S. Ren, and J. Sun, ''Deep residual learning for image recognition,'' in Proc. IEEE Conf. Comput. Vis. Pattern Recognit., Jun. 2016, pp. 770–778.

[15] G. Huang, Z. Liu, L. V. D. Maaten, and K. Q. Weinberger, "Densely connected convolutional networks," in Proc. IEEE Comput. Vis. Pattern Recognit., 2017, pp. 4700–4708.

[16] B. Cui, X. -M. Dong, Q. Zhan, J. Peng and W. Sun, "LiteDepthwiseNet: A Lightweight Network for Hyperspectral Image Classification," in IEEE Trans. Geosci. Remote Sens., doi: 10.1109/TGRS.2021.3062372.

[17] H. Gao, Y. Yang, C. Li, L. Gao, and B. Zhang, "Multiscale residual network with mixed depthwise convolution for hyperspectral image classification,"IEEE Trans. Geosci. Remote Sens., vol. 59, no. 4, pp. 3396–3408, doi:10.1109/TGRS.2020.3008286.

[18] F. Chollet, ''Xception: Deep learning with depthwise separable convolutions,'' in Proc. IEEE Conf. Comput. Vis. Pattern Recognit. (CVPR), Jul. 2017, pp. 1251–1258.

[19] X. Li, D. Song and Y. Dong, "Hierarchical Feature Fusion Network for Salient Object Detection," in IEEE Trans. Image Process., vol. 29, pp. 9165-9175, Sep. 2020.

[20] L.-C. Chen, G. Papandreou, I. Kokkinos, K. Murphy, and A. L. Yuille, "Semantic image segmentation with deep convolutional nets and fully connected CRFs," 2014, arXiv:1412.7062. [Online]. Available: http://arxiv.org/abs/1412.7062

[21] J. Cao et al., "DO-Conv: Depthwise Over-parameterized Convolutional Layer". [Online]. Available: https://arxiv.org/abs/2006.12030

[22] S. K. Roy, G. Krishna, S. R. Dubey and B. B. Chaudhuri, "HybridSN: Exploring 3-D–2-D CNN Feature Hierarchy for Hyperspectral Image Classification," in IEEE Geosci. Remote Sens Lett., vol. 17, no. 2, pp. 277-281, Feb. 2020.